\documentclass{article}

\usepackage{microtype}
\usepackage{graphicx}
\usepackage{subfigure}
\usepackage{booktabs} 
\usepackage{float}
\usepackage{lipsum}
\usepackage{comment}
\usepackage{courier}
\usepackage{stackengine}%
\usepackage{amsmath}
\usepackage{lipsum} 
\usepackage{kotex}
\usepackage{comment}

\usepackage{hyperref}

\usepackage[accepted]{icml2024}

\usepackage{amsmath}
\usepackage{amssymb}
\usepackage{mathtools}
\usepackage{amsthm}
\newcommand{\ie}{\textit{i}.\textit{e}., }
\newcommand{\eg}{\textit{e}.\textit{g}., }

\usepackage[capitalize,noabbrev]{cleveref}

\theoremstyle{plain}

\theoremstyle{definition}

\theoremstyle{remark}

\usepackage[textsize=tiny]{todonotes}

\icmltitlerunning{FRAG: Frequency Adapting Group for Diffusion Video Editing}

\begin{document}

\twocolumn[
\icmltitle{FRAG: Frequency Adapting Group for Diffusion Video Editing}



\icmlsetsymbol{equal}{*}

\begin{icmlauthorlist}
\icmlauthor{Sunjae Yoon}{kaist}
\icmlauthor{Gwanhyeong Koo}{kaist}
\icmlauthor{Geonwoo Kim}{kaist}
\icmlauthor{Chang D. Yoo}{kaist}
\end{icmlauthorlist}

\icmlaffiliation{kaist}{Korea Advanced Institute of Science
and Technology (KAIST)}

\icmlcorrespondingauthor{Chang D. Yoo}{cd\_yoo@kaist.ac.kr}

\icmlkeywords{Diffusion model, Text-based video editing, Frequency analysis}

\vskip 0.3in
]



\printAffiliationsAndNotice{}  

\begin{abstract}
In video editing, the hallmark of a quality edit lies in its consistent and unobtrusive adjustment. Modification, when integrated, must be smooth and subtle, preserving the natural flow and aligning seamlessly with the original vision. Therefore, our primary focus is on overcoming the current challenges in high quality edit to ensure that each edit enhances the final product without disrupting its intended essence. However, quality deterioration such as blurring and flickering is routinely observed in recent diffusion video editing systems. We confirm that this deterioration often stems from high-frequency leak: the diffusion model fails to accurately synthesize high-frequency components during denoising process. To this end, we devise Frequency Adapting Group (FRAG) which enhances the video quality in terms of consistency and fidelity by introducing a novel receptive field branch to preserve high-frequency components during the denoising process. FRAG is performed in a model-agnostic manner without additional training and validates the effectiveness on video editing benchmarks (i.e., TGVE, DAVIS). The project is available at: \href{https://dbstjswo505.github.io/FRAG-page/}{\texttt{dbstjswo505.github.io/FRAG-page}}
\end{abstract}

\section{Introduction}
\label{Introduction}
\begin{figure}[t]
\begin{center}
\centerline{\includegraphics[width=\columnwidth]{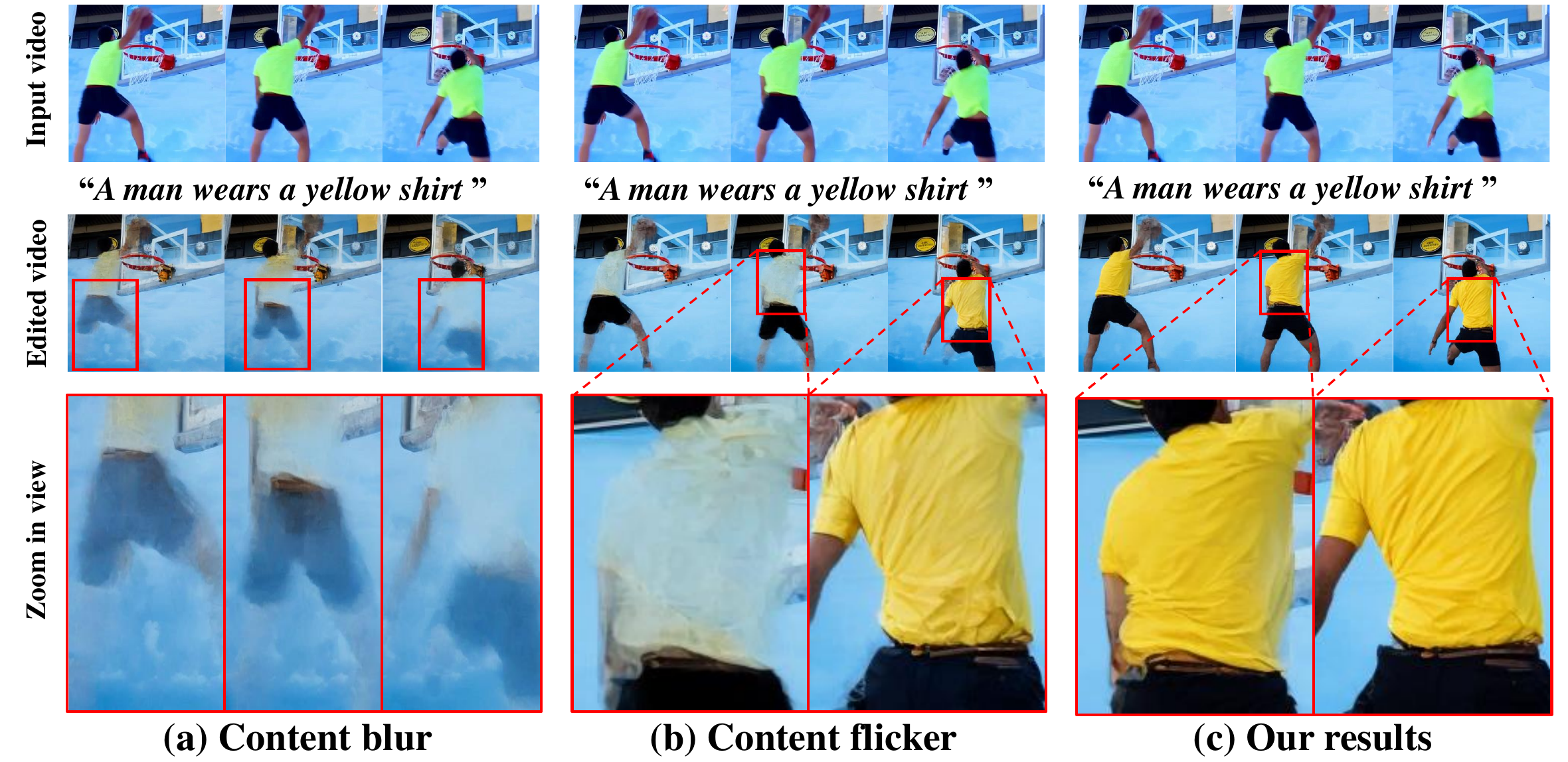}}
\caption{Illustration of video quality deterioration represented into two distinct categories: (a) content blur and (b) content flicker. For the comparison, we present our results in (c).}
\label{introduction}
\end{center}
\vskip -0.3in
\end{figure}
\begin{figure}[h!]
\begin{center}
\centerline{\includegraphics[width=\columnwidth]{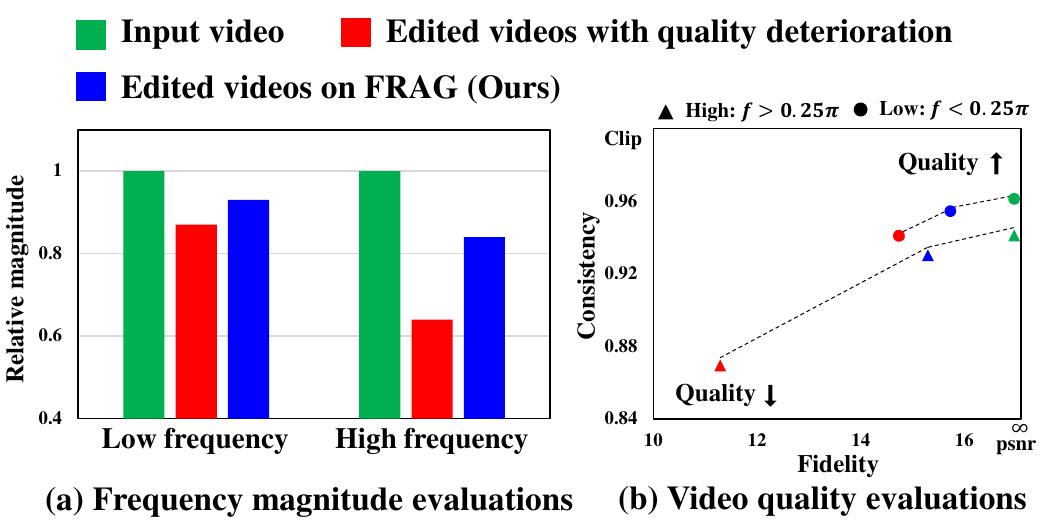}}
\caption{(a) Frequency magnitude evaluations of videos according to low and high frequencies. (b) Video quality evaluations about frame consistency and fidelity according to low and high-frequency components in videos. Normalized frequency $0<f<\pi$, low frequency:  $f< 0.25 \pi$, high frequency: $f> 0.25 \pi$.}
\label{introduction_b}
\end{center}
\vskip -0.4in
\end{figure}
Denoising diffusion models \cite{dhariwal2021diffusion,song2020score,song2020denoising,ho2020denoising} have significantly advanced the generative capabilities of artificial intelligence, leading to groundbreaking achievements in image, speech, and video generation.
We focus here on video editing based on diffusion which holds immense promise for revolutionizing the entertainment industry.
Video editing systems \cite{bar2022text2live,wu2023tune,geyer2023tokenflow} are designed to work with both the input video and a target text prompt that outlines the user's desired modifications. The systems incorporate these modifications into the video, ensuring that the edits are seamless and unobtrusive. 
%
This process is then carefully managed to produce a coherent final output that maintains a natural flow, whilst aligning closely with the original input video.

\begin{figure*}[t!]
\begin{center}
\centerline{\includegraphics[width=\textwidth]{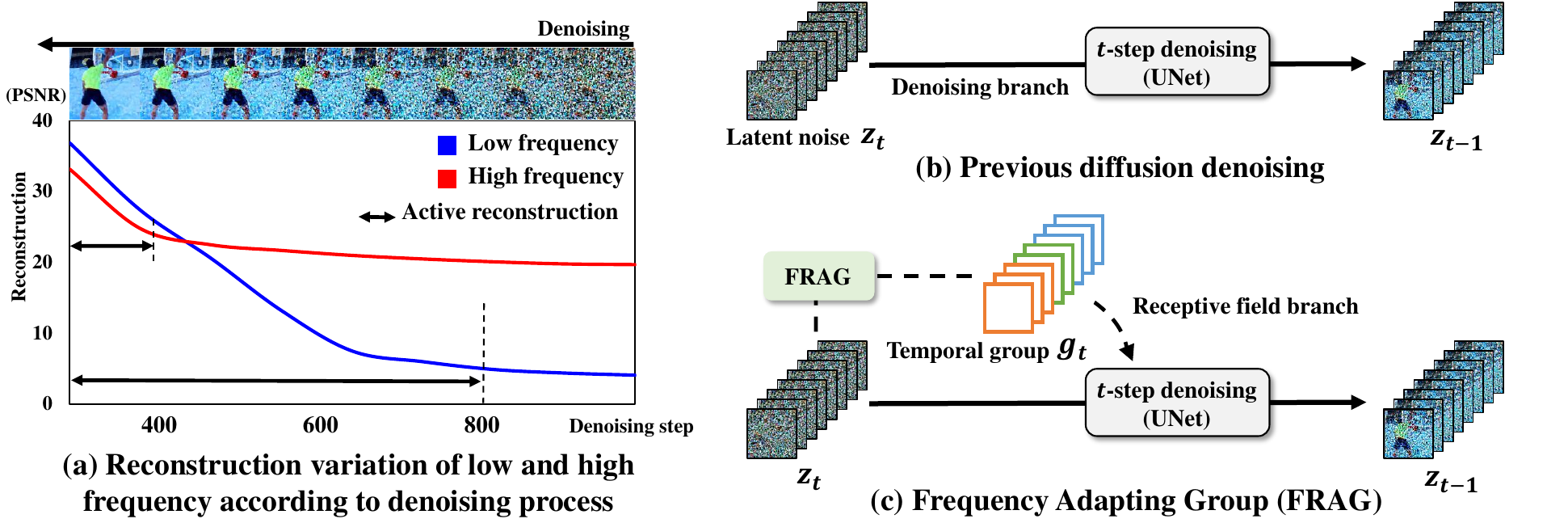}}
\caption{(a) shows experimental observations about latent noise reconstruction in terms of low and high frequencies, where high-frequency components are synthesized later in denoising than low frequencies. (b) illustrates previous video diffusion denoising and (c) illustrates our proposed denoising with the receptive field branch using Frequency Adapting Group (FRAG) to enhance the quality of editing.}
\label{introduction2}
\end{center}
\vskip -0.35in
\end{figure*}
Recent advancements \cite{wu2023tune,liu2023video,geyer2023tokenflow} in video editing systems have aimed at preserving the temporal consistency across edited frames. 
However, a significant challenge persists as these systems often struggle with maintaining the quality of various attributes, including the color and shape of objects. 
This inconsistency manifests not just over time but also across the spatial dimensions of the video, leading to a deterioration in the overall quality of the edits.
To be specific, Figure \ref{introduction} illustrates two distinct types of quality deterioration: (1) content blur and (2) content flicker.
As shown in Figure \ref{introduction} (a), the content blur denotes that attributes (\eg color and shape) synthesized for a content (\eg shirt) are irregularly mixed with other unintended contents (\eg background) in the entire video, which makes the content faint or unclear.
Figure \ref{introduction} (b) also shows another case of quality deterioration as a content flicker.
This indicates a disruption of visual continuity in a synthesized attribute at a certain moment, causing the attribute (\eg shirt color) to display contrasting characteristics (\eg light and dark) at different times.

Our observation suggests that blurring and flickering are due to a high-frequency leak in the diffusion denoising process.
The high-frequency leak denotes a shortfall of the video diffusion model's ability to accurately synthesize high-frequency attributes during the denoising process, leading to the lack of high-frequency components.
Figure \ref{introduction_b} presents experimental evidence about the high-frequency leak of current diffusion video editing systems \cite{geyer2023tokenflow,wu2023tune,khachatryan2023text2video}.
In Figure \ref{introduction_b} (a), we collected edited videos exhibiting quality deterioration and measured the average magnitude of frequency by converting them into low and high frequencies using a spatial frequency filter.
These videos show a deficiency in high-frequency components.
Furthermore, in Figure \ref{introduction_b} (b), qualities related to high frequency such as frame consistency and fidelity are degraded compared to those of the input video.\footnote{For consistency, due to the lack of supervision of editing, we applied editing (\eg style transfer) conforming to the consistency patterns in the input video, and measured clip score of input and output videos. For fidelity, we measure peak signal-to-noise (psnr) about the unedited region between input and output videos. Please note the difference of psnr between low and high frequencies in the samples (\ie red circle and triangle) exhibiting the deterioration.}
We further investigate to identify denoising dynamics in latent noise in terms of low and high frequency.
Figure \ref{introduction2} (a) shows the reconstruction\footnote{We measure the psnr between the input video and video decoded by each step latent noise.} of low and high-frequency latent noises according to the denoising step.
Notably, low-frequency components are reconstructed in the early denoising steps, while high frequencies tend to be reconstructed later.
This denotes that it is crucial to properly capture and preserve synthesized attributes at each frequency generation.

To achieve this, we devise Frequency Adpating Group (FRAG) for diffusion video editing, which enhances the quality of edited videos by effectively preserving high-frequency components.
As shown in Figure \ref{introduction2} (c), FRAG has an auxiliary branch for denoising process on top of the original denoising branch in Figure \ref{introduction2} (b).
This branch is defined as receptive field branch which guides denoising UNet \cite{ronneberger2015u} to properly synthesize the frequency components during the denoising process.
To be specific, this receptive field decides the frame-level operating range for the quality enhancement modules (\eg attention, propagation) within the UNet.
Previously, this field has employed fixed sliding windows or the entire video length \cite{geyer2023tokenflow, wu2023tune}, where both inevitably lead to a high-frequency leak problem.\footnote{A fixed small receptive field leads to flicker and a wide field leads to blur. Please see more details of this in Appendix \ref{fixed_receptive}.}
Thus, we devise a dynamic receptive field referred to as temporal group $g_{t}$, which adaptively refines the field 
according to synthesized frequency variations in each denoising step $t$.
Following the frequency characteristics of denoising in Figure \ref{introduction2} (a), $g_{t}$ builds large receptive fields in early denoising to facilitate the generation of low frequencies. 
As the denoising progresses, $g_{t}$ shifts to forming numerous smaller fields for high frequencies.
FRAG works in a model-agnostic manner without additional training and validates its effectiveness of quality on video editing benchmarks (\ie TGVE, DAVIS).

\section{Related Work}
\label{related_work}
\subsection{Diffusion-based Video Editing}
Video editing aims to edit the input video as seamlessly and unobtrusively as possible incorporating the target text descriptions.
The pre-trained text-to-image diffusion models \cite{rombach2022high,ramesh2022hierarchical} have presented an effective solution for generative editing, where earlier works \cite{kim2022diffusionclip,hertz2022prompt} in image editing laid the foundation for the development of controlled synthesis of visual information.
Extending the work in image, diffusion-based video editing \cite{molad2023dreamix,wu2023tune} has been attempted based on the video diffusion models \cite{ho2022video,hong2022cogvideo}.
To achieve accurate editing outcomes aligned with the target text, it is crucial to have controlled synthesis capabilities.
Thus, there have been lines of works \cite{zhang2023adding,liu2023video} to improve text-conditioned editing controllability. 
To enhance the efficiency of diffusion editing, zero-shot frameworks \cite{khachatryan2023text2video,qi2023fatezero} have been proposed.
These frameworks remove the process of training with the input video.
In particular, maintaining video quality across resulting frames is another crucial issue for video editing. 
We further elaborate on this in the following.
\subsection{Diffusion Video Editing Quality Enhancement}
The quality of edited video is evaluated through two standards: (1) frame consistency and (2) fidelity.
Frame consistency refers to the uniformity and coherence of consecutive frames in a video, while fidelity refers to the degree to which the edited video maintains the integrity and quality of the original content that is not meant to be altered.
There are three popular choices for diffusion video editing to video quality enhancement: (1) attention, (2) propagation, and (3) prior guidance.
Attention-based approach \cite{wu2023tune,liu2023video} is a method to highlight the visual commonality across frames based on their feature similarities.
It is effective in maintaining consistency based on contextual understanding of video scenes.
The propagation-based approach \cite{khachatryan2023text2video,geyer2023tokenflow} selects a pivotal key frame and shares its visual attribute with the attributes in frames within a given temporal receptive field of propagation. 
This ensures a highly consistent video at a visual attribute level.
The prior guidance methods \cite{cong2023flatten,chai2023stablevideo} perform editing following precomputed prior observations (\eg optical flow, object mask), which effectively enhances fidelity to the input video.
Although all of these approaches have pursued quality enhancement, they are still vulnerable to quality deterioration due to high-frequency leaks. 
Thus, we design an adaptive receptive field branch to guide quality enhancement modules to have robustness on the frequency variation in denoising.

\section{Preliminaries}
\label{sec:prelim}
\subsection{Denoising Diffusion Probabilistic Models}
\label{gen_inst}
Denoising diffusion probabilistic models (DDPMs) \cite{ho2020denoising} are parameterized Markov chains to sequentially reconstruct a noisy data $\{x_{1}$,$\cdots$, $x_{T}\}$ based on initial raw data $x_{0}$.
To construct this, Gaussian noise is gradually added upto $x_{T}$ via the Markov transition $q(x_{t}|x_{t-1}) = \mathcal{N}(x_{t};\sqrt{\alpha_{t}}x_{t-1},(1-\alpha_{t})I)$ utilizing a pre-defined schedule $\alpha_{t}$ across steps $t\in\{1,\cdots,T\}$.
This procedure is termed as \textit{forward process} in the diffusion model.
The counterpart to this, known as the \textit{reverse process}, involves the diffusion model estimating $q(x_{t-1}|x_{t})$ through trainable Gaussian transitions $p_{\theta}(x_{t-1}|x_{t}) = \mathcal{N}(x_{t-1};\mu_{\theta}(x_{t},t),\sigma_{\theta}(x_{t},t))$, beginning from the normal distribution $p(x_{T}) = \mathcal{N}(x_{T};0,I)$.
The training objective of diffusion model is to maximize log-likelihood $\textrm{log}(p_{\theta}(x_{0}))$ updating parameters $\theta$.
To this, we can apply variational inference of maximizing the variational lower bound about $\textrm{log}(p_{\theta}(x_{
0}))$, which builds a closed form of KL divergence\footnote{See the detailed proof in Appendix \ref{proof}.} between two distributions $p_{\theta}$ and $q$.
This whole process is summarized as introducing a denoising network $\epsilon_{\theta}(x_{t}, t)$ to predict noise $\epsilon \sim \mathcal{N}(0,I)$ as given below:
\begin{equation}
\begin{aligned}
\mathbb{E}_{x,\epsilon \sim \mathcal{N}(0,1),t \sim \mathcal{U}\{1,T\}}[||\epsilon - \epsilon_{\theta}(x_{t},t)||_{2}^{2}].
\end{aligned}
\end{equation}
where $\mathcal{U}\{1,T\}$ is discrete uniform distribution between 1 and $T$ for training robustness on each step $t$.
\subsection{Denoising Diffusion Implicit Model}
Denoising diffusion implicit model (DDIM) \citep{song2020denoising} accelerates the reverse process of DDPM, which samples noisy data with a smaller number of $T$ as below: 
\begin{equation}
\begin{aligned}
x_{t-1} = \sqrt{\frac{\alpha_{t-1}}{\alpha_{t}}}x_{t} + \left( \sqrt{\frac{1}{\alpha_{t-1}}-1} - \sqrt{\frac{1}{\alpha_{t}}-1} \right) \epsilon
\end{aligned}
\end{equation}
We can also inverse this process to compute latent noise as $x_{t+1} = \sqrt{\frac{\alpha_{t+1}}{\alpha_{t}}}x_{t} + \left( \sqrt{\frac{1}{\alpha_{t+1}} - 1} - \sqrt{\frac{1}{\alpha_{t}} - 1} \right) \epsilon$, referred to as DDIM inversion process. 
For diffusion editing, noise initialization with this enhances fidelity to the input video.
\subsection{Text-conditioned Diffusion Model}
\label{sec:3.2}
The text-conditioned diffusion model reconstructs the output data $x_{0}$ from random noise conditioned on a text prompt $\mathcal{T}$.
The training objective also incorporates text condition under latent space for semantic interaction as $\mathbb{E}_{z,\epsilon,t}[||\epsilon - \epsilon_{\theta}(z_{t},t,\mathbf{c})||_{2}^{2}]$, where $z_{t} = E(x_{t})$ is a latent noise encoding (\eg VQ-VAE \cite{van2017neural}) and $\mathbf{c}=\psi(\mathcal{T})$ is textual embedding (\eg CLIP \cite{radford2021learning}).
Diffusion video editing takes $z_{t}$ as the encoding of video data, and $\mathbf{c}$ for encoding the target text prompt.
\begin{figure*}[t]
\begin{center}
\centerline{\includegraphics[width=\textwidth]{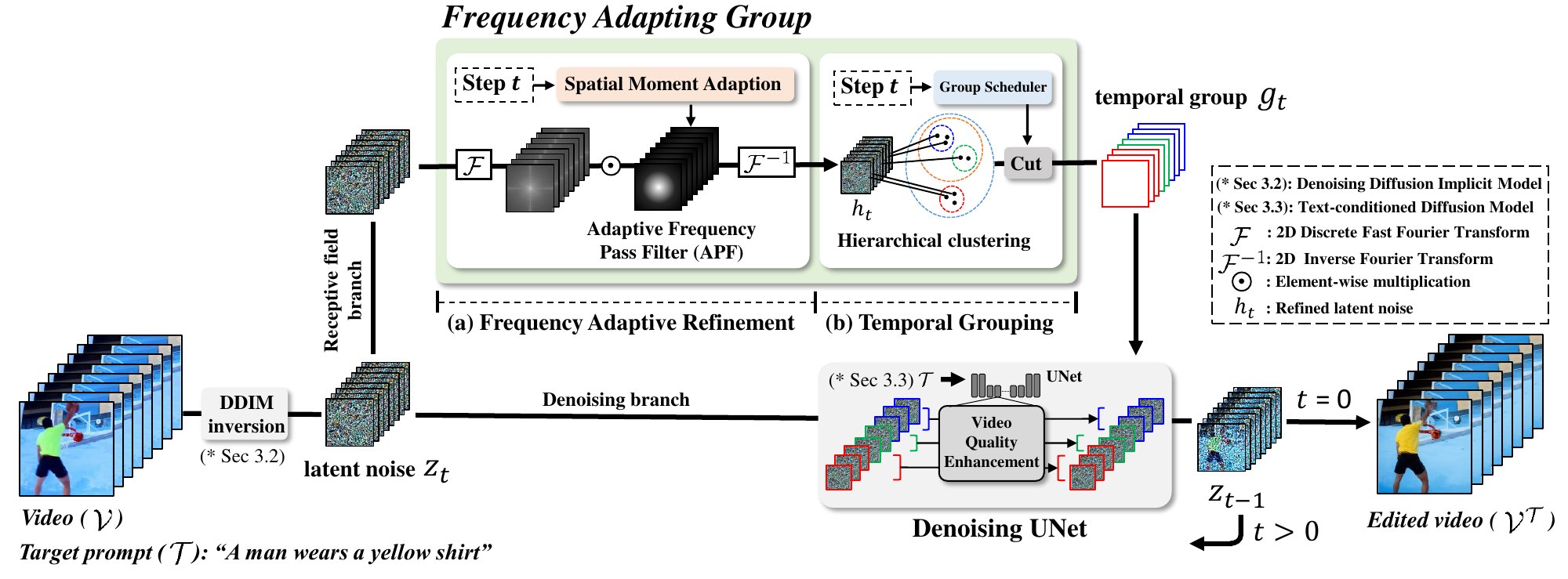}}
\caption{Illustration of Frequency Adapting Group (FRAG). FRAG takes $t$ step latent noise $z_{t}$ and produces receptive field $g_{t}$ referred to as temporal group. The $g_{t}$ guides denoising UNet to adaptively synthesize the frequency components according to frequency variations of latent noise during the denoising process. FRAG contains (a) frequency adaptive refinement that enhances the visual quality of attributes within latent noise and (b) temporal grouping that clusters latent noise frames to build $g_{t}$.}
\label{model}
\end{center}
\vskip -0.25in
\end{figure*}
\subsection{Denoising Spectral Characteristic}
Frequency analysis of latent space revealed the process of denoising is patterned by the spatial frequencies within the latent noises.
For clarity, we term this pattern as `denoising spectral characteristics'. 
This concept encapsulates the sequential synthesis of spatial frequency of latent noise in the denoising process: Low frequency is synthesized in the early stages, followed by the synthesis of high frequency in the subsequent phases.
\begin{figure}[t]
\begin{center}
\centerline{\includegraphics[width=\columnwidth]{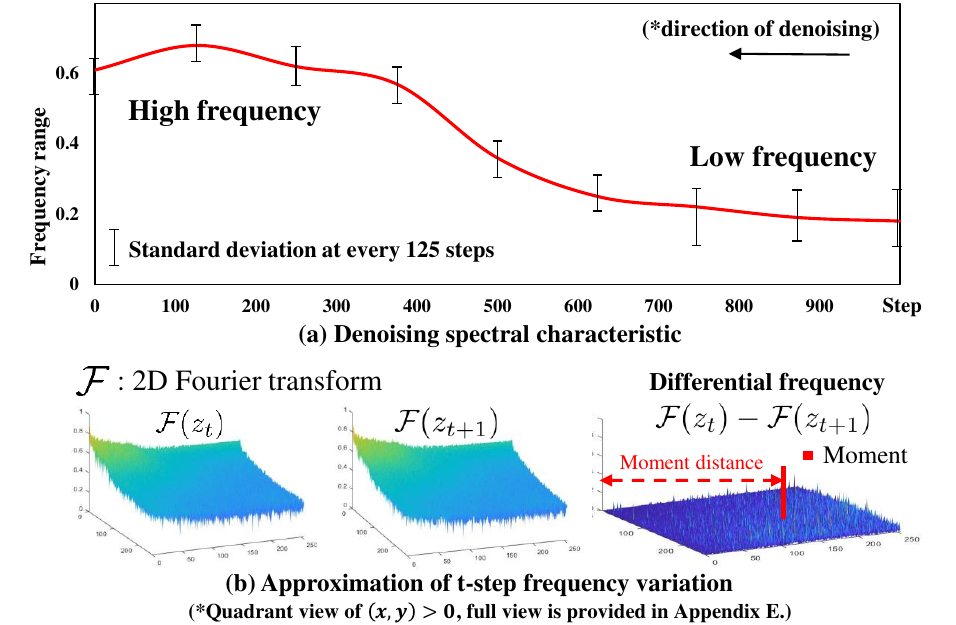}}
\caption{(a) Denoising spectral characteristic: Average frequency variation according to denoising from 1000 to 0 step on 800 videos in TGVE \cite{wu2023cvpr} and UCF-101 \cite{soomro2012ucf101}. (b) $t$ step frequency variation: It is approximated by a normalized distance of moment in the differential frequency distribution.}
\label{dsc}
\end{center}
\vskip -0.4in
\end{figure}
Figure \ref{dsc} (a) demonstrates experimental observations of the denoising spectral characteristics, which measures the frequency variation of synthesized latent noise according to denoising.
This shows a gradual increase in the frequency as the denoising process advances.
To estimate the frequency variation, in Figure \ref{dsc} (b), we transform each latent noise into spatial frequency and calculate each step differential frequency distribution by subtracting the previous step frequency.
Based on differential distribution, we measure a normalized distance of the moment in it (Please see details in Sec 4.1) and approximate the distance as frequency variation.
Leveraging this spectral characteristic, we present the Frequency Adapting Group in the following.
\section{Frequency Adapting Group}
Diffusion video editing system takes inputs of video $\mathcal{V}$ and target prompt $\mathcal{T}$, where it produces edited video $\mathcal{V}^{\mathcal{T}}$ conforming to the meaning of $\mathcal{T}$.
Figure \ref{model} shows the application of Frequency Adapting Group (FRAG) into the general diffusion video editing system, which allows both supervised (\ie tuning) and unsupervised (\ie tuning-free) models.
FRAG aims to enhance the quality of edited videos by effectively preserving high-frequency components.
At each denoising step $t$, FRAG takes an input latent noise $z_{t}$ and produces a receptive field $g_{t}$ referred to as a temporal group.
This temporal group guides the quality enhancement module (\eg attention, propagation) in denoising UNet to preserve the frequencies dynamically synthesized during the denoising process.
To perform this, FRAG comprises two main modules: (1) Frequency Adaptive Refinement (Sec 4.1) and (2) Temporal Grouping (Sec 4.2).
Frequency adaptive refinement refines the visual quality of synthesized attributes within latent noise by applying our adaptive frequency pass filter.
Based on this refinement, temporal grouping clusters frames with similar latent noise into temporal groups based on shared content.
Finally, these groups are provided as receptive fields for quality enhancement of denoising UNet. 
\begin{figure}[t]
\begin{center}
\centerline{\includegraphics[width=\columnwidth]{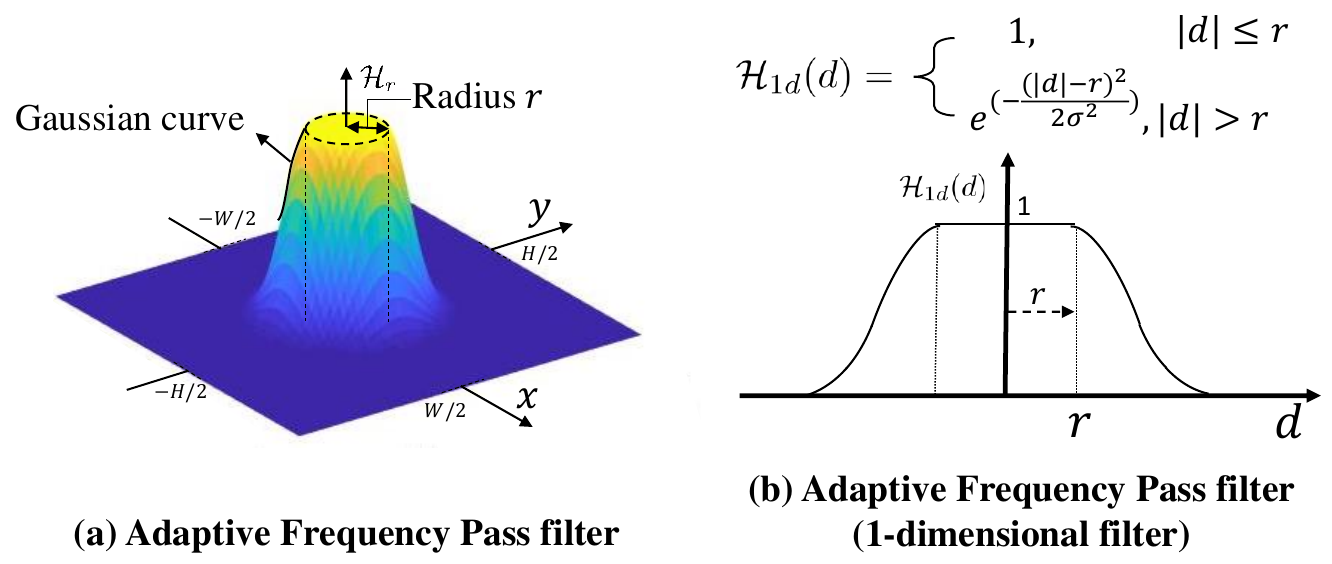}}
\caption{Visualization of Adaptive Frequency Pass filter (APF). (a) shows APF in 3D view, where $x, y$ are spatial axes and $\mathcal{H}_{r}$ is the filter value. (b) shows a 1-dimensional APF $\mathcal{H}_{1d}$ of 2D view.}
\label{apf}
\end{center}
\vskip -0.5in
\end{figure}
\subsection{Frequency Adaptive Refinement}
Frequency Adaptive Refinement aims to enhance the visual quality of synthesized attributes within the latent noise of each step.
According to the denoising spectral characteristic, attributes are progressively synthesized from low to high frequency, such that we have devised an adaptive frequency pass filter (APF) capable of progressively passing the frequency ranging from low to high corresponding to the currently synthesized frequency. 
Thus, APF reveals the synthesized attributes better within latent noise by keeping track of frequency synthesizing trends.
To be specific, as shown in Figure \ref{apf} (a), the APF is a 2-dimensional low-frequency pass filter denoting $\mathcal{H}_{r} \in \mathbb{R}^{L \times W \times H}$, where $W, H$ are width and height of the filter. The $L$ is the number of filters corresponding to frame length.
It is a cylindrical structure of radius $r$ (\ie $0<r<\sqrt{(W/2)^2 + (H/2)^2}$) whose edges follow a Gaussian curve for smoothing effect (Figure \ref{apf} (b) presents a 1-dimensional APF to enhance the understanding of its shape).
Formally, the $\mathcal{H}_r$ can be defined as below:
\begin{equation}
\mathcal{H}_{r}^{i} = 
    \begin{cases} 
		1 &  d \leq r \\ 
         e^{-(d-r)^2/2\sigma^{2}} &  d > r ,
    \end{cases}
\end{equation}
where $\sigma$ is coefficient for scaling the Gaussian curve. The superscript $i$ denotes the $i$-th filter. (We omit this in the following for the simplicity.) The $d$ is a distance of each point ($x,y$) in 2D frequency domain from the center point, satisfying $d(x,y) = \sqrt{x^2 + y^2}$.
We multiply this $\mathcal{H}_{r}$ into the latent noise frequency and convert it back to the real domain, which preserves the components inside the radius $r$ of the latent noise frequency as given below:
\begin{equation}
h_{t} = \mathcal{F}^{-1}(\mathcal{H}_{r} \odot \mathcal{F}(z_{t})) \in \mathbb{R}^{L \times W \times H \times C},
\end{equation}
where $z_{t}$ is $t$ step latent noise and $C$ is the channel.
$\mathcal{F},\mathcal{F}^{-1}$ are discrete time fast Fourier transform and inverse transform. $\odot$ is element-wise multiplication with broadcasting\footnote{Since in discrete time, $\mathcal{F}(z_{t}) \in \mathbb{R}^{L \times W \times H \times C}$ builds same dimension of latent noise $z_{t} \in \mathbb{R}^{L \times W \times H \times C}$. Thus, spatial frequency filter $\mathcal{H}_{r} \in \mathbb{R}^{L \times W \times H}$ is broadcasting to the channel axis.}.

Therefore, $h_{t}$ is refined latent noise by $\mathcal{H}_{r}$.
By expanding $r$, the $\mathcal{H}_{r}$ encompasses the generated frequency.\footnote{High frequencies can also be distinguished as they are much lower than the Gaussian noise frequencies within the latent noise.}
To achieve this, we introduce a spatial moment adaption below.

\paragraph{Spatial Moment Adaption.}
The spatial moment adaption adjusts the radius $r$ of the adaptive filter $\mathcal{H}_{r}$ to include the frequency generated at each denoising step.
%
To identify the generated frequency at $t$ step, as shown in Figure \ref{dsc} (b), we obtain differential frequency $\mathcal{Z}_{t}$ by subtracting the previous step frequency, from the $t$ step as $\mathcal{Z}_{t} = \mathcal{F}(z_{t}) - \mathcal{F}(z_{t+1})$.
Thus, the $\mathcal{Z}_{t}$ contains spatial frequencies generated during the $t$ step denoising process, to cover these frequencies by radius $r$ in APF, we calculate a point of the spatial moments about $\mathcal{Z}_{t}$ as $M_{x}, M_{y}$ on a space $(x,y)>0$ satisfying below:
\begin{equation}
\begin{aligned}
M_{x} = \frac{\sum_{x,y} x \mathcal{Z}_{t}(x,y)}{\sum_{x,y}\mathcal{Z}_{t}(x,y)}, M_{y} = \frac{\sum_{x,y}y \mathcal{Z}_{t}(x,y)}{\sum_{x,y}\mathcal{Z}_{t}(x,y)}.
\end{aligned}
\end{equation}
Finally, the radius for $\mathcal{H}_{r}$ is defined as $r = d(M_{x}, M_{y}) + d_{0}$ with margin $d_{0}$.
Therefore $r$ is adaptively updated following the $t$ step synthesized frequency distribution.
\begin{figure}[h]
\begin{center}
\centerline{\includegraphics[width=\columnwidth]{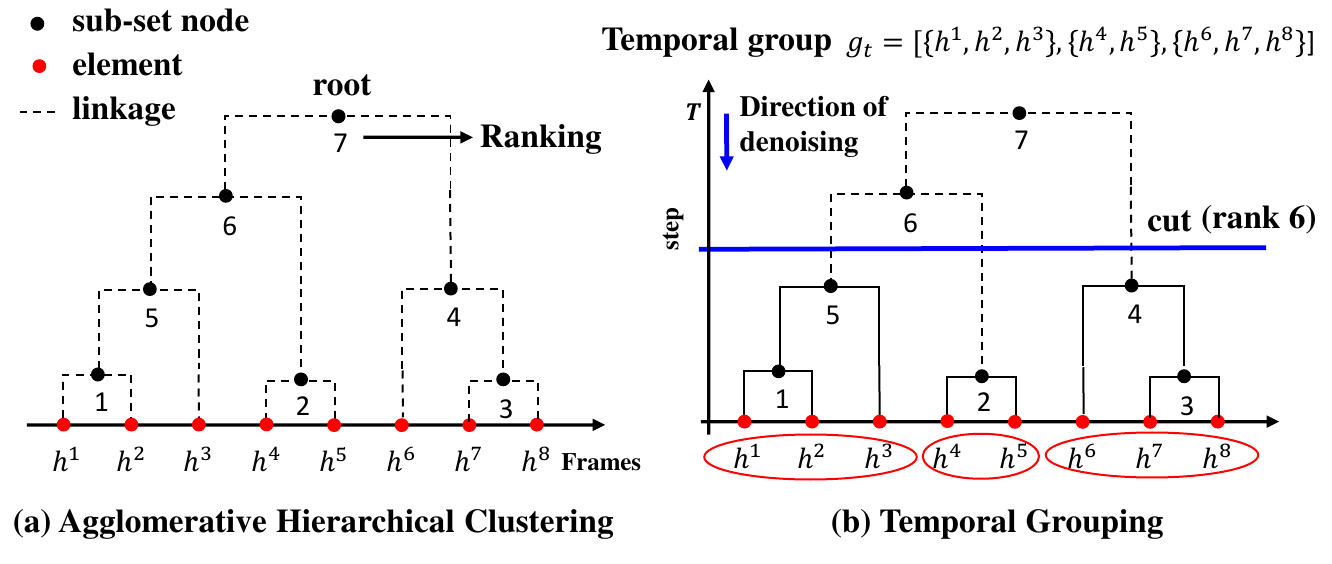}}
\caption{(a) shows the agglomerative hierarchical clustering and (b) shows temporal grouping by group scheduler to cut the tree (a).}
\label{hc}
\end{center}
\vskip -0.3in
\end{figure}
\subsection{Temporal Grouping}
Temporal grouping aims to provide a receptive field composed of multiple frame groups, where each group contains latent noise features containing similar attributes.
Based on the refined latent noise $h_{t} \in \mathbb{R}^{L \times W \times H \times C}$, the temporal grouping measures their frame distances between consecutive frames and clusters frames based on the distances, producing multiple frame groups $g_{t}$.
We first define frame distance between $i$-th and $j$-th frame latent noises by applying Euclidean distance as given below:
\begin{equation}
D(h^{i},h^{j}) = ||h^{i} - h^{j}|| \in \mathbb{R},
\end{equation}
where $h^{i}$ is $i$-th refined latent noise feature.\footnote{we skip subscript step $t$ for the simplicity.}
Based on our defined frame distance, we apply hierarchical clustering (HC) \cite{jain1988algorithms} with agglomeration (\ie bottom-up).
As shown in Figure \ref{hc} (a), the HC constructs a binary merge tree, initiating with individual elements (\ie single frame latent noise, $h^{i}$). 
It progressively merges the nearest sub-sets (\ie multi frames) in pairs, advancing towards the root of the tree which ultimately encompasses all elements.
For each frame to cluster the closest pair of sub-sets, we extend the frame distance $D(h^{i},h^{j})$ into a sub-set distance between any two sub-sets of frames $X$ and $Y$ as $\Delta(X,Y) = \textrm{min}_{h^{i} \in X, h^{j} \in Y}D(h^{i},h^{j})$ with minimum distance linkage.
Therefore, the HC algorithm\footnote{Please refer algorithm of agglomerative HC in Appendix \ref{AHC_alg}.} builds a tree that links each $h^{i}$ in order of minimum distance about $\Delta(X,Y)$, where the number below the linkage in Figure \ref{hc} (a) denotes the ranking of linkages.
To build temporal groups $g_{t}$ using this tree, as shown in Figure \ref{hc} (b), we cut one of the linkages in the tree.
For the selection of linkage to cut, we design a `group scheduler' which selects the ranking of linkage from high to low according to denoising.
This is because, according to the denoising spectral characteristic, the attributes of high-frequency components are synthesized in the latter step of denoising (\eg $t <$ 400), so the receptive field also needs to be narrow for appropriately clustering high-frequency components such as fine-grained attributes.
Thus, the group scheduler cuts off high rankings in the beginning of denoising to form a small number of temporal groups with a wide range and cuts out low rankings in the later stages to form a large number of temporal groups with a short range.
Formally, we choose a logistic curve for the group scheduler to stably decide the number of rankings according to each step $t$ as given below:
\begin{equation}
n_{\textrm{cut}} = \left \lceil{n_{\textrm{root}} \times (\alpha \textrm{log}(T - t) + 1)}\right \rceil,
\end{equation}
where $n_{\textrm{cut}}, n_{\textrm{root}}$ are the integer numbers of rank to cut and the root rank. 
$T=1000$ is a maximum step and $\alpha=-1/\textrm{log}(T-1)$ is scaler to fit output range from 1 to $n_{\textrm{root}}$.
After cutting, we construct a temporal group (\eg $g_{t}=[\{h^{1},h^{2},h^{3}\},\{h^{4},h^{5}\},\{h^{6},h^{7},h^{8}\}]$) by remained subsets.
\subsection{Plug-and-Play FRAG}
We integrate temporal group $g_{t}$ into video editing systems by applying it into a quality enhancement module (\eg attention, propagation) of video diffusion UNet. 
In general, when the quality module is defined as $f:\mathbb{R}^{l \times d} \rightarrow \mathbb{R}^{l \times d}$, $l$ is the range of frames (\eg $l = L$) to be performed of enhancement and $d$ is the feature dimension (\eg $d = W \times H \times C$).
We can simply update the $l$ as $g_{t}$ as below:
\begin{equation}
f: \mathbb{R}^{g_{t}^{i} \times d} \rightarrow \mathbb{R}^{g_{t}^{i} \times d},
\end{equation}
where the $g_{t}^{i}$ is the $i$-th group of temporal group.
Therefore, the diffusion model with FRAG adaptively denoises latent noise within the temporal groups designed for preserving synthesized frequencies throughout the denoising process.
\section{Experiment}
\subsection{Experimental Settings}
\paragraph{Implementation Details.}
Diffusion video editing systems that apply FRAG use CLIP \cite{radford2021learning} for the text encoder and VQ-VAE \cite{van2017neural} for the patch-wise image frames encoder.
We use a pre-trained Stable Diffusion v2.1 \cite{rombach2022high} for knowledge of editing.
The experimental settings are $W = H =64, L=48, C=4, \sigma=0.25, d_{0}=6$ on NVIDIA A100 GPU.
More details are also available in Appendix \ref{det}.
\paragraph{Data and Baseline.}
We validated videos using the TGVE and DAVIS datasets, both of which are video editing challenge datasets\footnote{https://sites.google.com/view/loveucvpr23/track4} containing 32 to 128 frames each.
%
FRAG is validated on recent editing systems including TokenFlow \cite{geyer2023tokenflow}, FLATTEN \cite{cong2023flatten}, Tune-A-Video (TAV) \cite{wu2023tune}, FateZero \cite{qi2023fatezero} on their public codes and papers.
\subsection{Evaluation Metrics}

Editing is measured based on the following five qualities: (1) frame consistency, (2), fidelity to input video, (3) spectral analysis, (4) textual alignment, (5) human preference.
The frame consistency measures image CLIP scores between sequential frames and measures Fréchet Video Distance (FVD) to evaluate the naturalness of videos.
The fidelity measures the preservation of original content in the unedited region using Peak Signal-to-Noise Ratio (PSNR) and Structural Similarity Index Measure (SSIM).
The spectral analysis measures consistency and fidelity in terms of low and high frequency.
The textual alignment assesses semantic coherence between a target prompt and an edited video, utilizing CLIP score.
For human preference, we analyze the preferences for edited videos based on the target prompt.
The details about capturing unedited regions for fidelity and human evaluation are provided in Appendix \ref{det}.
\begin{figure*}[t!]
\centering
    \includegraphics[width=1.0\textwidth]{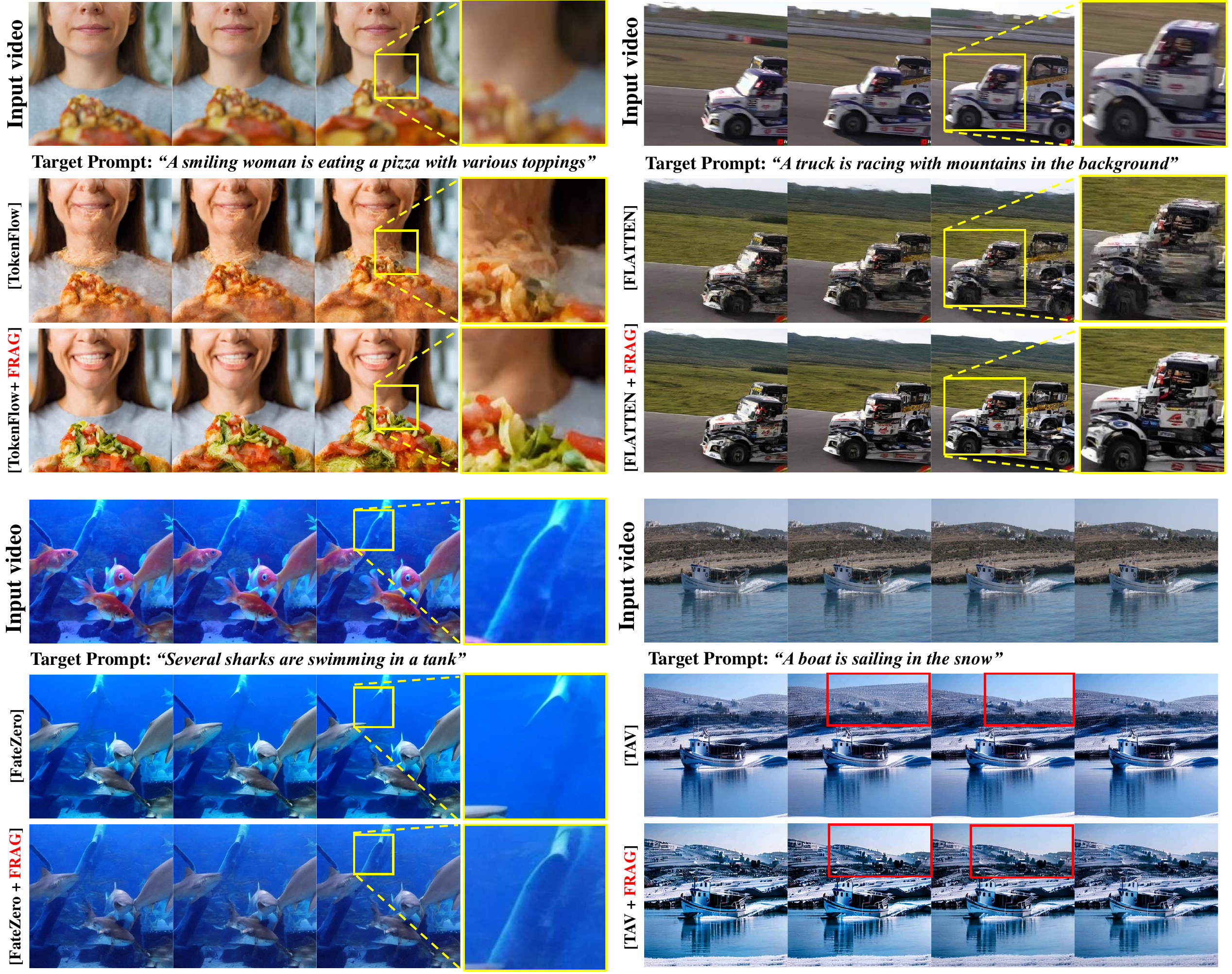}
   \caption{Qualitative result about applying FRAG on recent editing systems (TokenFlow: propagation-based model, FLATTEN: optical flow guidance, FateZero: zero-shot model, Tune-A-Video: attention based model), (Red box: content flicker, yellow box: content blur).}
\label{fig:qual}
\end{figure*}
\subsection{Experimental Results}
\paragraph{Qualitative Comparisons.}
To demonstrate the effectiveness of our proposed FRAG, as shown in Figure \ref{fig:qual}, we apply FRAG to the recent four video editing systems (\ie TokenFlow, FLATTEN, FateZero, TAV) and qualitatively evaluate the edit results.
The top left shows results pertaining to TokenFlow which relies on a fixed-size receptive field for quality enhancement module (\ie propagation) in all denoising steps.
This results in a content blur (\ie yellow box) in terms of the colors and edges.
However, TokenFlow with FRAG solves this blurring by dynamically configuring the receptive field during each denoising process, and it can be seen that the attributes become clearer.
The frames at the top right show the results about FLATTEN\footnote{We reproduce it based on their paper.}, where it uses auxiliary guidance (\ie optical flow) to preserve consistency using pre-trained model \cite{teed2020raft}.
The editing outcomes demonstrate high consistency and fidelity, yet they still reveal instances of content blur (\ie yellow box).
The model, when integrated with FRAG, effectively mitigates the blur, delivering clearer attributes while preserving a high degree of fidelity to the input video.
The frames at the bottom left show results about FateZero, which performs video editing in zero-shot approach.
In this model, blurring is also observed in the editing results.
The application of FRAG notably enhanced the blurring, concurrently, improving the fidelity (\ie Tree of background correctly is preserved).
We think that building a receptive field under consideration of frequency in latent noise can lead to consistent alterations of the same attribute during the attribute synthesizing process, ensuring uniformity in changes.
The frames displayed on the lower right are the outcomes of TAV, a tuning-based attention approach that has served as a foundational baseline for numerous editing systems.
TAV employs sliding window attention for quality enhancement, however, it exhibits severe temporal flickering.
When integrated with FRAG, TAV improves this flickering issue and preserves high frequencies such as fine-grained details.
\begin{table*}[h]
\caption{Quantitative results of edited videos on DAVIS and TGVE based on editing systems with FRAG about consistency (frame consistency), fidelity (fidelity to input video), spectral analysis (consistency and fidelity of low/high normalized frequency $f$, low: $f<$0.25$\pi$, high: $f>$0.25$\pi$), alignment (textual alignment), and human (preference). $\textrm{CLIP}^{\star}$: text-video clip,
$\textrm{CLIP}^{\dagger}$: image-image clip.}
\centering
\footnotesize
\begin{center}
\begin{tabular}{lcccccccc} \toprule[1pt]
& \multicolumn{2}{c}{\bf Consistency} &\multicolumn{2}{c}{\bf Fidelity} 
& \multicolumn{2}{c}{\bf Spectral Analysis}
&\multicolumn{1}{c}{\bf Alignment} & {\bf Human}
\\ \cline{2-9}
& $\textrm{CLIP}^{\dagger}$ ↑ & FVD ↓ & PSNR ↑ & SSIM ↑ & $\textrm{CLIP}^{\dagger}$ ↑ & PSNR ↑ & $\textrm{CLIP}^{\star}$ ↑ &Preference ↑ \\ \midrule
TAV          &0.932 &3452 &13.7 &0.647 & 0.961 / 0.872 &14.1 / 11.0 &25.2 &0.27 \\
TAV + FRAG   &\textbf{0.951} &\textbf{3251} &\textbf{14.9} &\textbf{0.687} & \textbf{0.965} / \textbf{0.913} &\textbf{15.2} / \textbf{13.2} &\textbf{25.7} &0.73 \\ \midrule 
FateZero  &0.945 &3241 &13.9 &0.651 & 0.969 / 0.893 & 14.3 / 12.6 & 24.5 &0.39 \\ 
FateZero + FRAG &\textbf{0.956} &\textbf{3119} &\textbf{15.3} &\textbf{0.694} & \textbf{0.971} / \textbf{0.911} &\textbf{15.6} / \textbf{13.9} &\textbf{25.1} &0.61 \\ \midrule
FLATTEN   &0.962 &3002 &14.2 &0.672 & 0.968 / 0.911 & 14.6 / 12.7 & 25.4 &0.41 \\
FLATTEN + FRAG          &\textbf{0.970} &\textbf{2951} &\textbf{15.3} &\textbf{0.702} &\textbf{0.971} / \textbf{0.928} & \textbf{16.0} / \textbf{14.8} &\textbf{25.6} &0.59 \\ \midrule
TokenFlow   &0.968 &2984 &15.1 &0.691 & 0.972 / 0.931 & 15.1 / 13.1 & 25.8 &0.43 \\
TokenFlow + FRAG          & \textbf{0.978} &\textbf{2841} &\textbf{18.2} &\textbf{0.736} &\textbf{0.981} / \textbf{0.954} &\textbf{18.2} / \textbf{16.3} & \textbf{26.4} &0.57 \\\bottomrule[1pt]
\end{tabular}
\end{center}
\label{mytab:1}
\end{table*}
\paragraph{Quantitative Results.}
Table \ref{mytab:1} presents evaluations of edited videos on DAVIS and TGVE of recent editing systems with FRAG in five criteria (\ie consistency, fidelity, spectral analysis, alignment, human evaluation). 
The baselines include different types of quality enhancement modules (\ie TAV: attention, TokenFlow: propagation, FLATTEN: optical flow), and the effectiveness of FRAG is confirmed in all the models. 
The consistency and fidelity are effectively enhanced in the models with FRAG.
In spectral analysis, we separate video into high-frequency and low-frequency components using a frequency filter, where FRAG significantly improves video quality of high frequency.
\begin{figure}[t]
\begin{center}
\centerline{\includegraphics[width=\columnwidth]{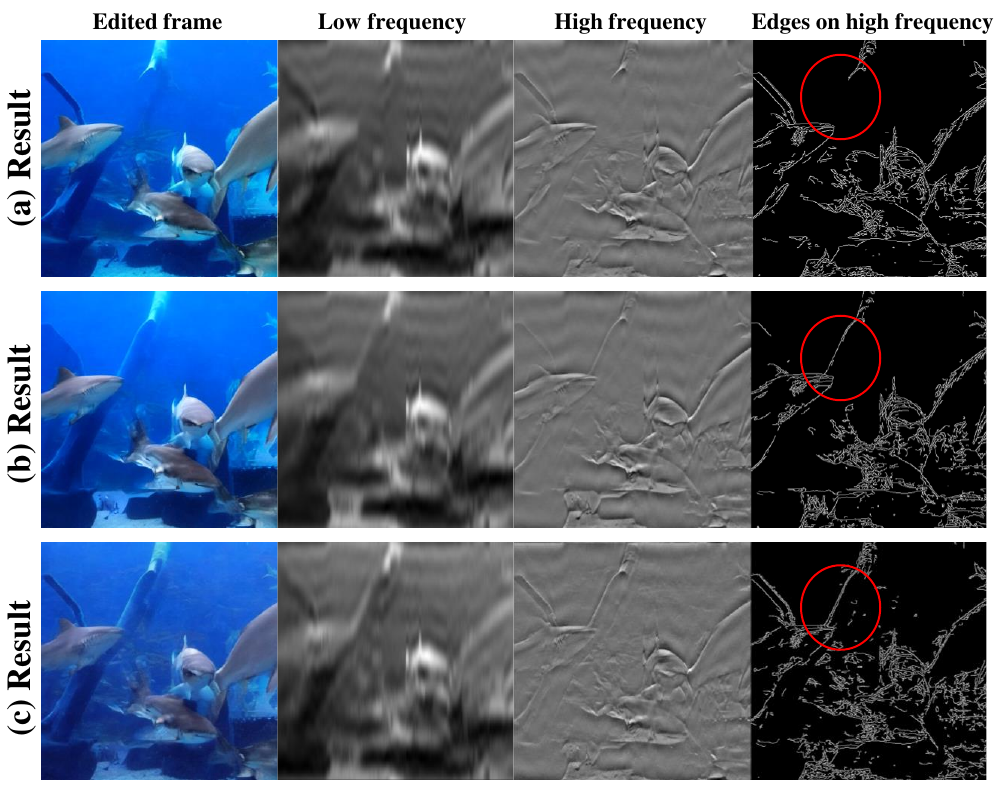}}
\caption{Ablation studies on main modules in FRAG about low and high frequency and edges on high frequency. (a): baseline (FateZero), (b): baseline + FRAG (temporal grouping), (c): baseline + FRAG (frequency adaptive refinement + temporal grouping). The input video is presented in Figure \ref{fig:qual}.}
\label{abs1}
\end{center}
\vskip -0.35in
\end{figure}
\begin{figure}[t]
\begin{center}
\centerline{\includegraphics[width=\columnwidth]{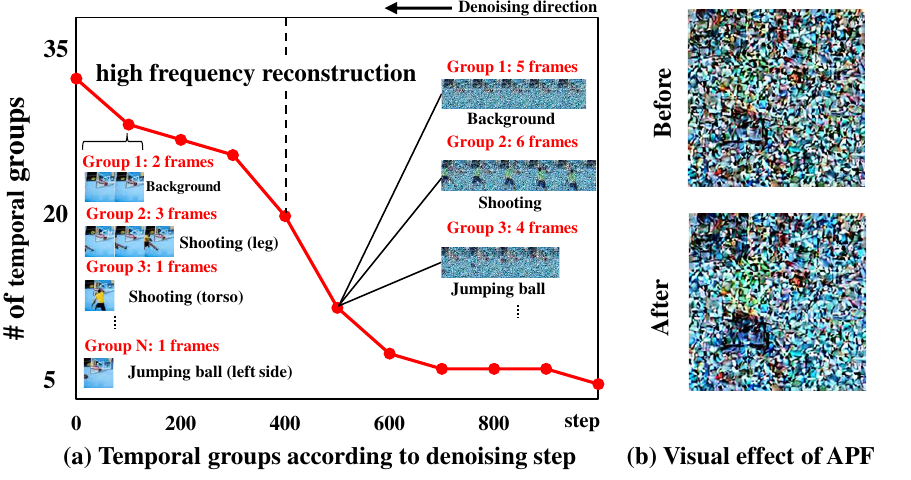}}
\caption{(a) shows variation in the number of temporal groups according to the denoising step. The groups are progressively abundant and fine-grained to contain high-frequency details. (b) shows visual effectiveness before/after applying the adaptive frequency pass filter (APF) on decoded latent noise at step 781.}
\label{abs2}
\end{center}
\vskip -0.35in
\end{figure}
\subsection{Ablation Study}
Figure \ref{abs1} presents ablative studies on FRAG in terms of low and high frequencies in edited videos.
The results (a) are the editing with baseline (FateZero), where quality deterioration, such as blurring due to mixing with trees in the background, is identified.
The results (c) confirm that combining FRAG with the model effectively mitigates this deterioration in both low and high frequencies.
Especially, frequency adaptive refinement plays a key role in this mitigation, improving the high-frequency details with more edges about the original shape.
Figure \ref{abs2} (a) shows variations in temporal groups during the denoising process.
At the start of denoising, there are fewer groups with many frames in each.
Interestingly, frames are grouped by similar scenes (\eg dunk-shoot scenes).
When high frequencies appear (\ie 400 step), group numbers surge, by clustering frames of similar fine details (\eg scene of the appearance of legs or torso).
Figure \ref{abs2} (b) qualitatively shows the effect of applying APF to latent noise, enhancing visual quality by extracting low-frequency attributes in Gaussian noise.
\section{Discussion}
\paragraph{Limitations.} Our model can be applied to several consistency modules but has a certain degree of sensitivity across modules.
Empirical studies show FRAG excels in propagation methods.
It's also effective with prior-guidance methods like optical flow, but less so compared to the former.
We consider this is because consistency is enforced by the guidance prior.
In this way, our proposed adjusting the receptive field also has sensitive effectiveness according to its application and there are still many additional improvements needed to enhance video quality comprehensively.
Moreover, although this paper employs frequency characteristics for temporal grouping and some works \cite{si2023freeu,huang2024fouriscale,he2024freestyle} are also concerned about the frequency for the diffusion model, it is also important to understand the scene characteristics of the image/video to achieve robustness even for videos with longer and more diverse scenes. 
To the best of our knowledge, employing scene knowledge for the diffusion model has never been studied before.
To this end, utilizing video search technology \cite{yoon2022selective,yoon2023counterfactual} appears promising by constructing scene-aware temporal grouping.
Additionally, employing recent weakly-supervised \cite{yoon2023scanet,ma2020vlanet} and unsupervised \cite{luo2024zero} methods can reduce the training resource and enhance the effectiveness of the approach.
\paragraph{Future work.}
Video editing has recently surged in popularity, yet numerous unresolved issues remain. 
We briefly introduce the various methods we are considering for future work to address the issues.
Video editing performance remains highly sensitive to prompts. 
To address this, it is essential to further incorporate prompt optimization and tuning methods \cite{yoon2023information}, similar to those used in image editing \cite{kawar2023imagic}.
Current video editing technology simply relies on human's intuitive decisions about the success of editing, but it needs to be integrated with more detailed automatic control of the editing effect.
For this purpose, integrating calibration systems \cite{yoon2023esd} seems novel for fine-grained controllability of editing. 
Furthermore, current video editing is slow.
The zero-shot \cite{geyer2023tokenflow,qi2023fatezero} can solve this but it offers limited editability.
The tuning method provides high editability but requires a significant amount of time. 
Thus, it is essential to enhance the video tuning method to address these challenges similar to the works \cite{koo2024wavelet} in the image.
Finally, there is still a lack of research on the vision-language model applying video editing/generation technology.
We believe that the convergence of generative technologies will bring explosive innovation to image/video high-level tasks including dialogue \cite{yoon2022information}, and commonsense reasoning \cite{liang2022visual}.
\section{Conclusion}
This paper proposes Frequency Adaptive Group (FRAG) which enhances the video quality of diffusion video editing systems in a model-agnostic manner.
We found the spectral characteristics of latent noise that low-frequency attributes emerge in the early stages, followed by the synthesis of higher-frequency attributes.
Based on this characteristic, FRAG enhances the video quality according to the frequency variation of synthesized latent noise.
\section*{Impact Statement}
Visual generative models are associated with ethical challenges, including the creation of unauthorized counterfeit content, risks to privacy, and issues of fairness.
Our work is built on these generative models, inheriting their vulnerabilities.
Therefore, it's crucial to tackle these issues through a combination of robust regulations and technical safeguards.
We think that researchers assume responsibility for these issues, actively working to implement technical safeguards.
We are also considering measures like adopting learning-based digital forensics and implementing digital watermarking.
These actions are designed to guide the ethical use of visual generative models, ensuring their responsible and positive application.
\section*{Acknowledgements}
This work was supported by Institute for Information \& communications Technology Planning \& Evaluation (IITP) grant funded by the Korea government(MSIT) (No. 2021-0-01381, Development of Causal AI through Video Understanding and Reinforcement Learning, and Its Applications to Real Environments) and partly supported by Institute of Information \& communications Technology Planning \& Evaluation (IITP) grant funded by the Korea government(MSIT) (No.2022-0-00184, Development and Study of AI Technologies to Inexpensively Conform to Evolving Policy on Ethics).
%
%
%

\nocite{langley00}

\bibliography{frag}
\bibliographystyle{icml2024}

\clearpage
\appendix
\section*{Appendix}
\begin{figure}[t]
\centering
    \includegraphics[width=1.0\linewidth]{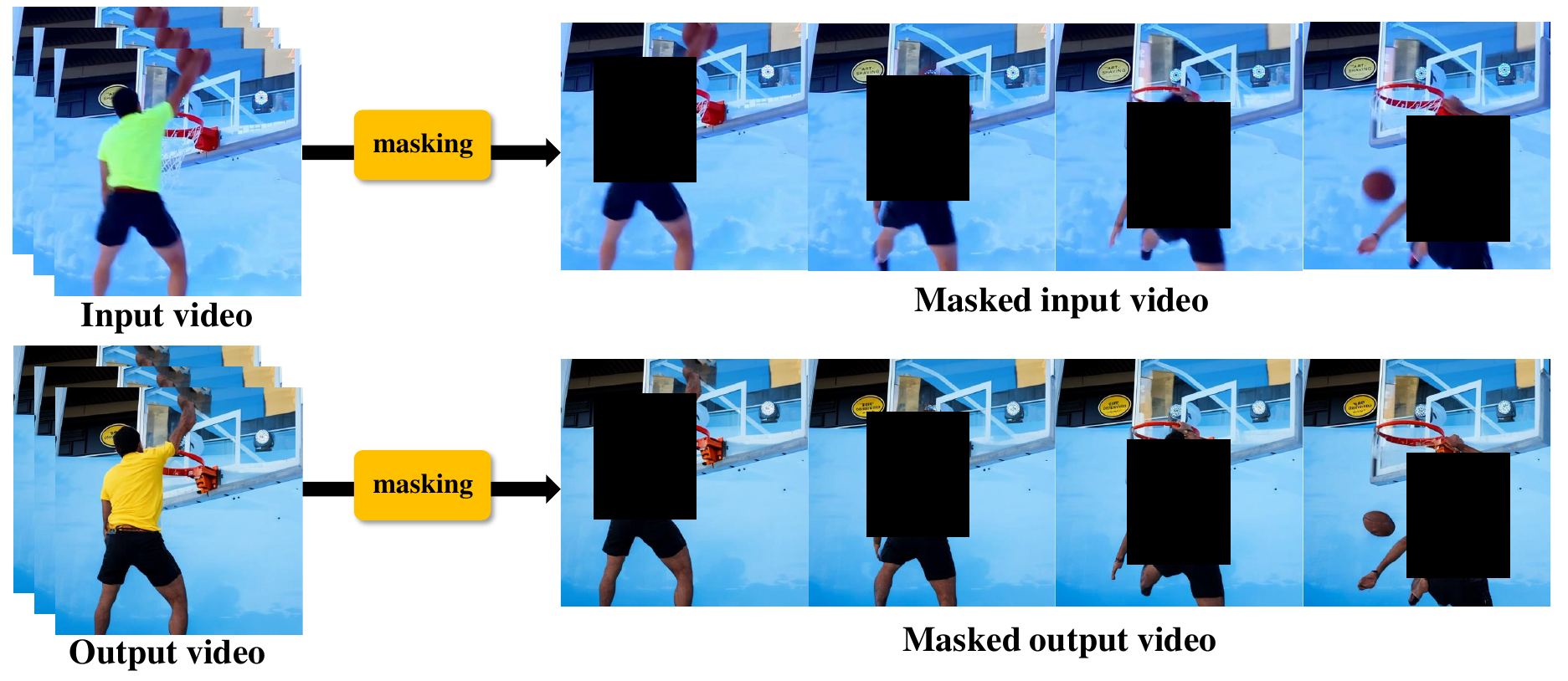}
   \caption{Masked input and output videos for fidelity metric assessment, applying the same masks in the edited regions.}
\label{fig:fidelity_masking_1}
\vskip -0.2in
\end{figure}
\section{Details of Implementation and Evaluations}
\label{det}
\paragraph{Implementation Details.}
For video encoding, all the baselines utilize VQ-VAE \cite{van2017neural}, which provides patch-wise encoding of each frame.
For text encoding the CLIP model (ViT-L/14) \cite{radford2021learning} is used for conditional text input.
For Equation (6), to compute frame distance we apply mean-pooing along the spatial domain and then perform Euclidean distance for the computational efficiency.
We set the margin $d_{0} = 6$ for the Sptial Moment Adaption, which was the most effective in our framework, where the Gaussian curve in Equation (3) also smoothly includes the frequency near the moment ($M_{x}$,$M_{y}$) in the 2-dimensional frequency domain.
We adjusted the minimum size of the temporal group along the frame axis to a range between 1 and 4, according to video quality enhancement modules (\eg propagation, attention).

\paragraph{Fidelity Evaluation Details.}
To assess fidelity to the input video, we applied the same zero mask to the edited areas in both the input and output videos, as illustrated in Figure \ref{fig:fidelity_masking_1}.
By masking the area for editing, we can evaluate the preservation of unedited content between the input and output videos, yielding PSNR and SSIM scores that reflect their similarity and consistency.
We utilize automatic detectors \cite{kirillov2023segment} to specify the areas of square mask for editing in both input and output videos.
For some edits (\ie background change) that are not proper by automatic detectors, we specify the edited region.
\paragraph{Human Evaluation Details.}
Human evaluation is conducted to assess preferences for the edited outcomes based on a specified target prompt.
Motivated the format of human evaluation in the work \cite{yoon-etal-2023-hear}, we conducted a survey comparing preferences for outputs from existing editing systems and the FRAG framework under consideration of semantic alignment, and video quality.
A survey was conducted with 36 participants from varied academic fields such as engineering, literature, and art.

\section{Closed Form of KL Divergence}
\label{proof}
The reverse process of DDPM is to approximate $q(x_{t-1}|x_{t})$ based on learnable Gaussian transitions $p_{\theta}(x_{t-1}|x_{t}) = \mathcal{N}(x_{t-1};\mu_{\theta}(x_t,t),\sigma_{\theta}(x_{t},t))$. 
We first define whole $T$ step transitions, by sequentially constructing them as $p_{\theta}(X) = p_{\theta}(x_{T})\prod_{t=1}^{T}p_{\theta}(x_{t-1}|x_{t})$.
where staring normal distribution $p(x_{T}) = \mathcal{N}(x_{T};0,\textit{I})$, this considers transitions of $X = x_{0:T}$.
For the training objective of $p_{\theta}(X)$, we should maximize log-likelihood $\textrm{log}(p_{\theta}(X))$. 
Otherwise, we can also apply variational inference by maximizing the variational lower bound $-\mathcal{L}_{vlb}$ as given below:
\begin{equation}
\begin{aligned}
-\mathcal{L}_{vlb} = \textrm{log}p_{\theta}(X) - D_{\textrm{KL}}(q(Z|X)||p_{\theta}(Z|X))\\
\leq \textrm{log}p_{\theta}(X),
\end{aligned}
\end{equation}
where $D_{\textrm{KL}}$ is the Kullback-Leibler divergence and the $Z$ is the latent variable using reparametrization trick by the variational auto-encoder.
The $q$ can be any distributions that we can approximate.
We inverse the inequality condition as $-\textrm{log}p_{\theta}(X) \leq \mathcal{L}_{vlb}$. 
Here, the $-\textrm{log}p_{\theta}(X)$ is conditioned by $\mathcal{L}_{vlb}$ by expanding it as $\mathcal{L}_{vlb} = \mathcal{L}_{T} + \mathcal{L}_{T-1} + \cdots + \mathcal{L}_{0}$, where they are defined with $1 \leq t \leq T$ as given below:
\begin{equation}
\begin{aligned}
&\mathcal{L}_{T} = D_{\textrm{KL}}(q(x_{T}|x_{0})||p_{\theta}(x_{T})),\\ 
&\mathcal{L}_{t} = D_{\textrm{KL}}(q(x_{t}|x_{t+1},x_{0})||p_{\theta}(x_{t}|x_{t+1})), \\ 
&\mathcal{L}_{0} = -\textrm{log}p_{\theta}(x_{0}|x_{1}).
\end{aligned}
\end{equation}
These terms make the closed form of KL divergence under step $t$ with a range of $0 \leq t \leq T$. 
%
\section{Agglomerative Hierarchical Clustering}
\label{AHC_alg}
\begin{algorithm}[h!]
   \caption{Agglomerative Hierarchical Clustering}
   \label{alg:example}
\begin{algorithmic}
   \STATE {\bfseries Input:} data $x_i$
   \STATE Initialize for each data element $x_{i} \in X$ its cluster singleton $G_{i} = \{x_{i}\}$ in a list
   \WHILE {there remain two elements in the list}
   \STATE Choose $G_{i}$ and $G_{j}$ so that $\Delta(G_{i},G_{j})$ is minimized among all pairs,
   \STATE Merge $G_{i,j} = G_{i} \cup G_{j}$,
   \STATE Add $G_{i,j}$ to the list,
   \STATE Remove $G_{i}$ and $G_{j}$ from the list.
    \ENDWHILE
    \STATE Return the remaining group in the list ($G_{\textrm{root}} = X$) as the dendrogram root.
\end{algorithmic}
\end{algorithm}
\begin{figure*}[t]
\centering
    \includegraphics[width=1.0\textwidth]{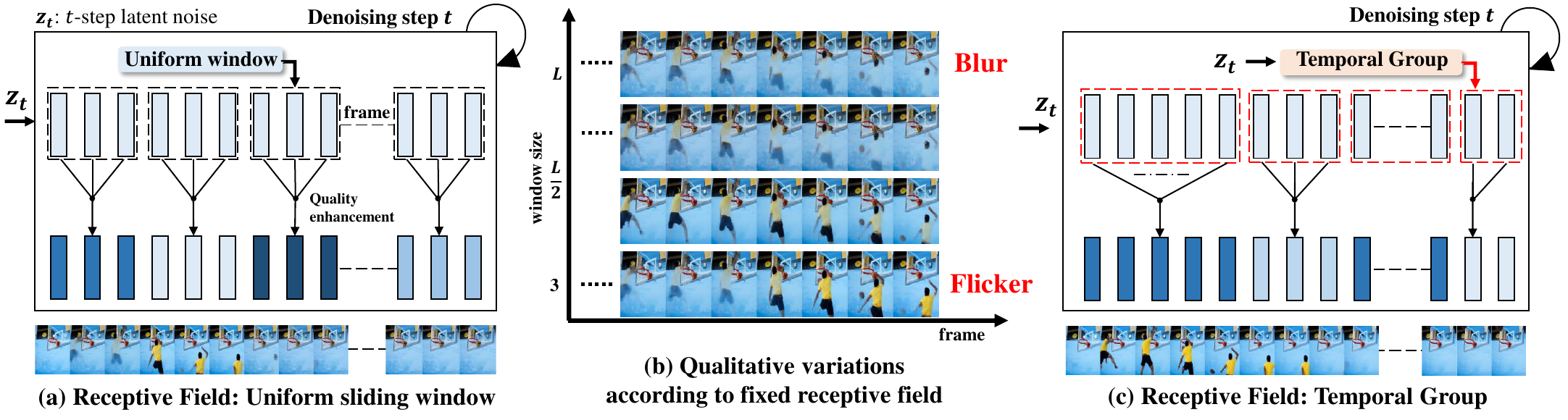}
   \caption{(a) Illustration of fixed receptive field using a uniform sliding window, (b) qualitative results about ablation study about video results according to window size. (c) Illustration of dynamic receptive field using temporal group.}
\label{fig:receptive_field}
\end{figure*}
\section{Ablation Studies on Fixed Receptive Field}
\label{fixed_receptive}
Figure \ref{fig:receptive_field} (a) presents a conceptual illustration of uniform sliding windows for receptive field for quality enhancement module (\eg temporal attention, propagation) of current systems \cite{wu2023tune,geyer2023tokenflow,liu2023video}.
Although this structure offers an intuitive understanding of the frame-level enhancement, it inevitably causes content blur or content flicker. 
As shown in Figure \ref{fig:receptive_field} (b), when applying the windows with a relatively large size (\eg frame length $L$), they make a blurred video.
latent noise interactions for quality enhancement in long-range receptive fields often overlook the high-frequency attributes synthesized in individual frames, leading to a blurring of content.
Conversely, if the receptive field is reduced (\ie window size of $2 \sim 4$), high-frequency components are generated, but these are ununiformly synthesized across each field.
Conversely, when the receptive field is reduced, high-frequency components are generated, but the low-frequency components are not positioned uniformly, so the high-frequency components generated above them are also not uniform.
Thus, this makes a content flicker.
Figure \ref{fig:receptive_field} (c) shows our proposed dynamic receptive field using a temporal group.
The temporal group (\ie red box) adaptively designs the receptive field according to the variational frequency synthesis during the denoising step.

\section{More Qualitative Results}
All the video samples are based on DAVIS, TGVE, and copyright-free videos at \hyperlink{here.}{https://www.pexels.com}.
Here we present qualitative results about (1) the frequency distribution of decoded latent noise, (2) more comparison results, and (3) more results with FRAG.
\paragraph{Frequency Distribution of Decoded Latent Noise.}
\begin{figure}[h!]
\centering
    \includegraphics[width=1.0\linewidth]{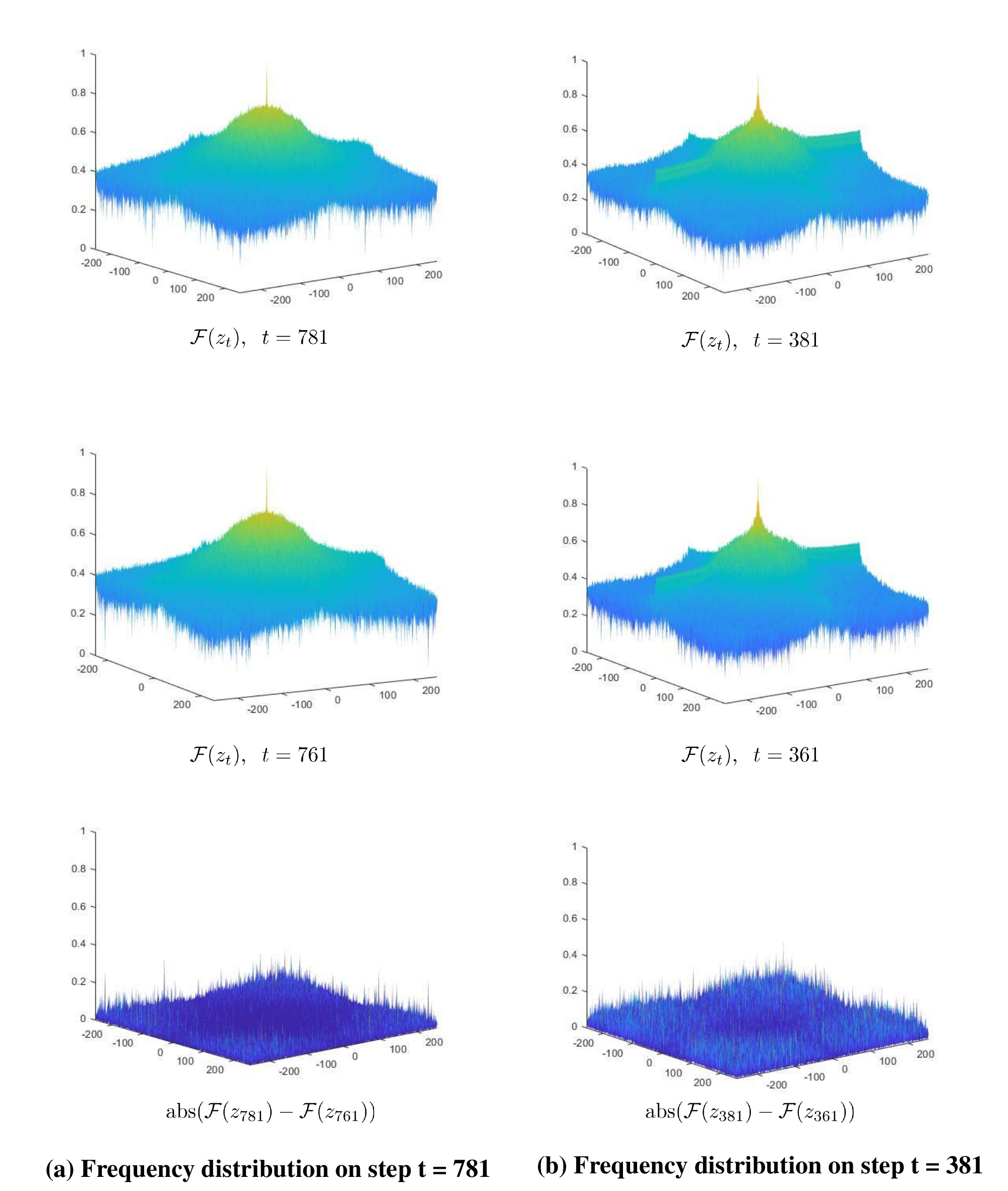}
   \caption{Illustration of a frequency distribution (full view on discrete frequency domain) on a decoded image about latent noise at step $t=781$, and $t=381$. (a) shows the differential frequency in the early stage of denoising (\ie $t=781$), where there is less information in the region of high frequency ($f>0.25 \pi$) in the difference. (b) shows the differential frequency in the latter stage of denoising (\ie $t=381$), where there is dense information in the high-frequency region. We used DDIM for denoising, such that the unit step is 20 for the sampling.}
\label{fig:app_freq}
\vskip -0.2in
\end{figure}
\paragraph{More Comparison Results.}
\begin{figure*}[b!]
\centering
    \includegraphics[width=1.0\textwidth]{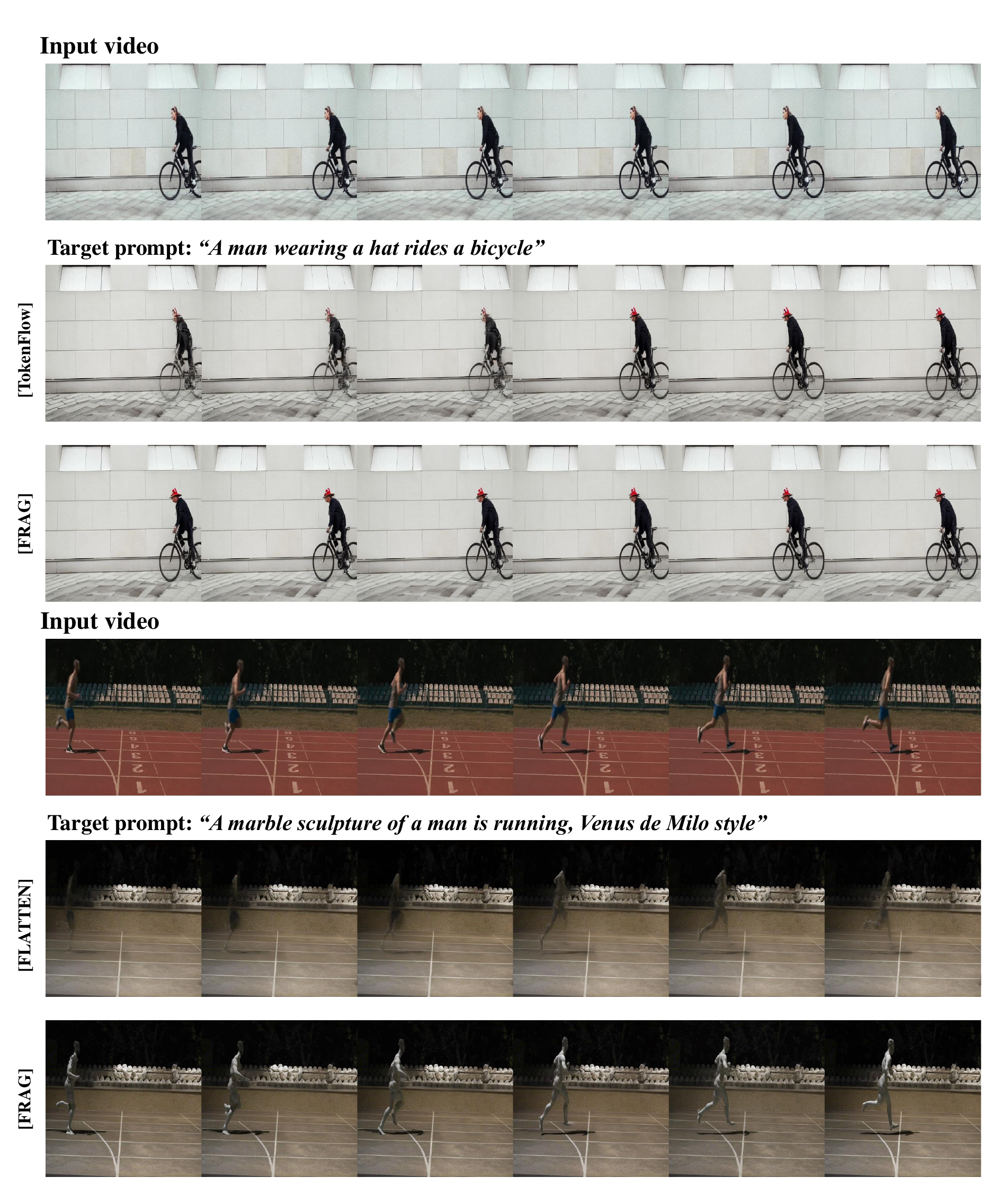}
\label{fig:page1}
\end{figure*}
\begin{figure*}[t]
\centering
    \includegraphics[width=1.0\textwidth]{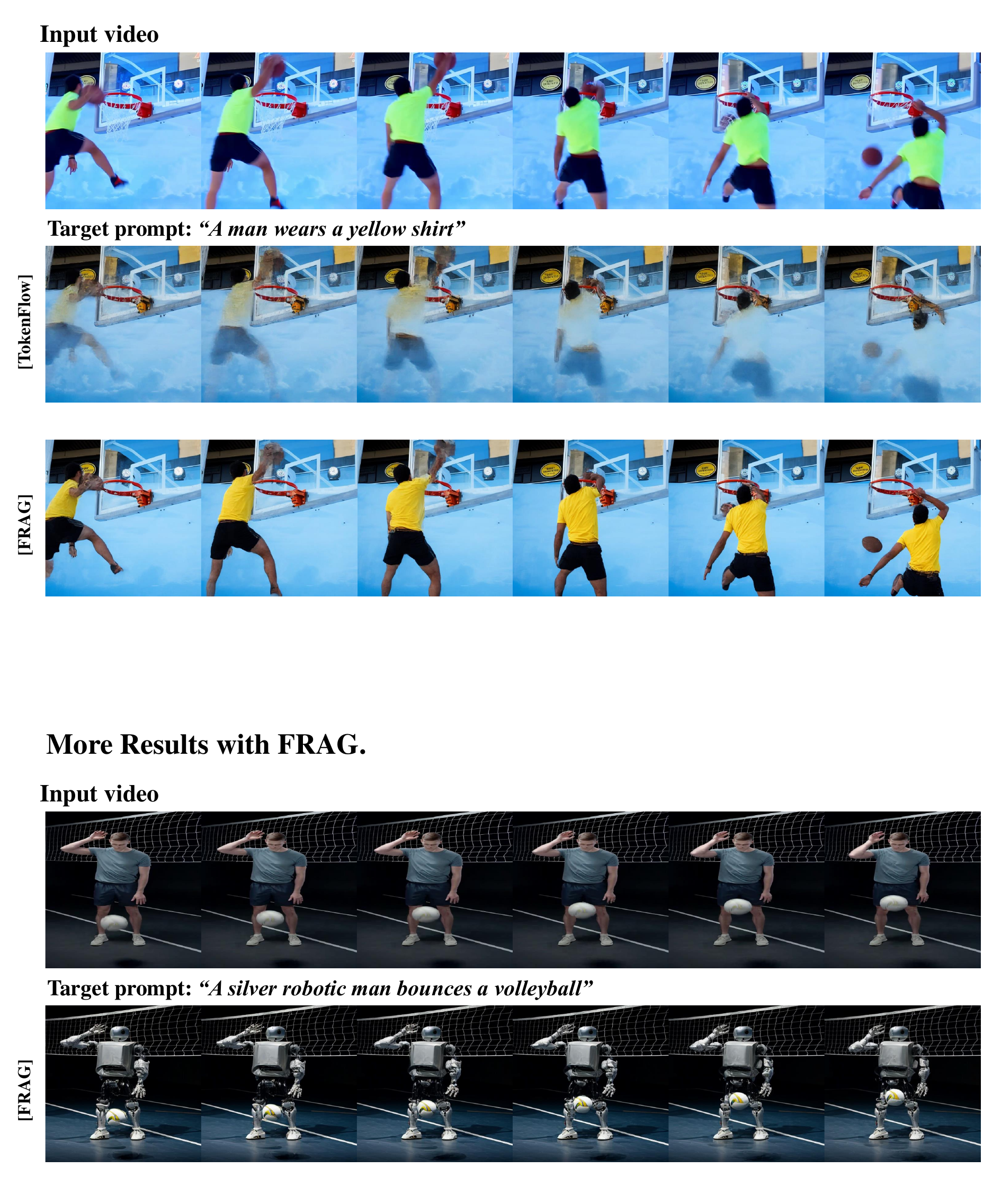}
\label{fig:page2}
\end{figure*}
\begin{figure*}[t]
\centering
    \includegraphics[width=1.0\textwidth]{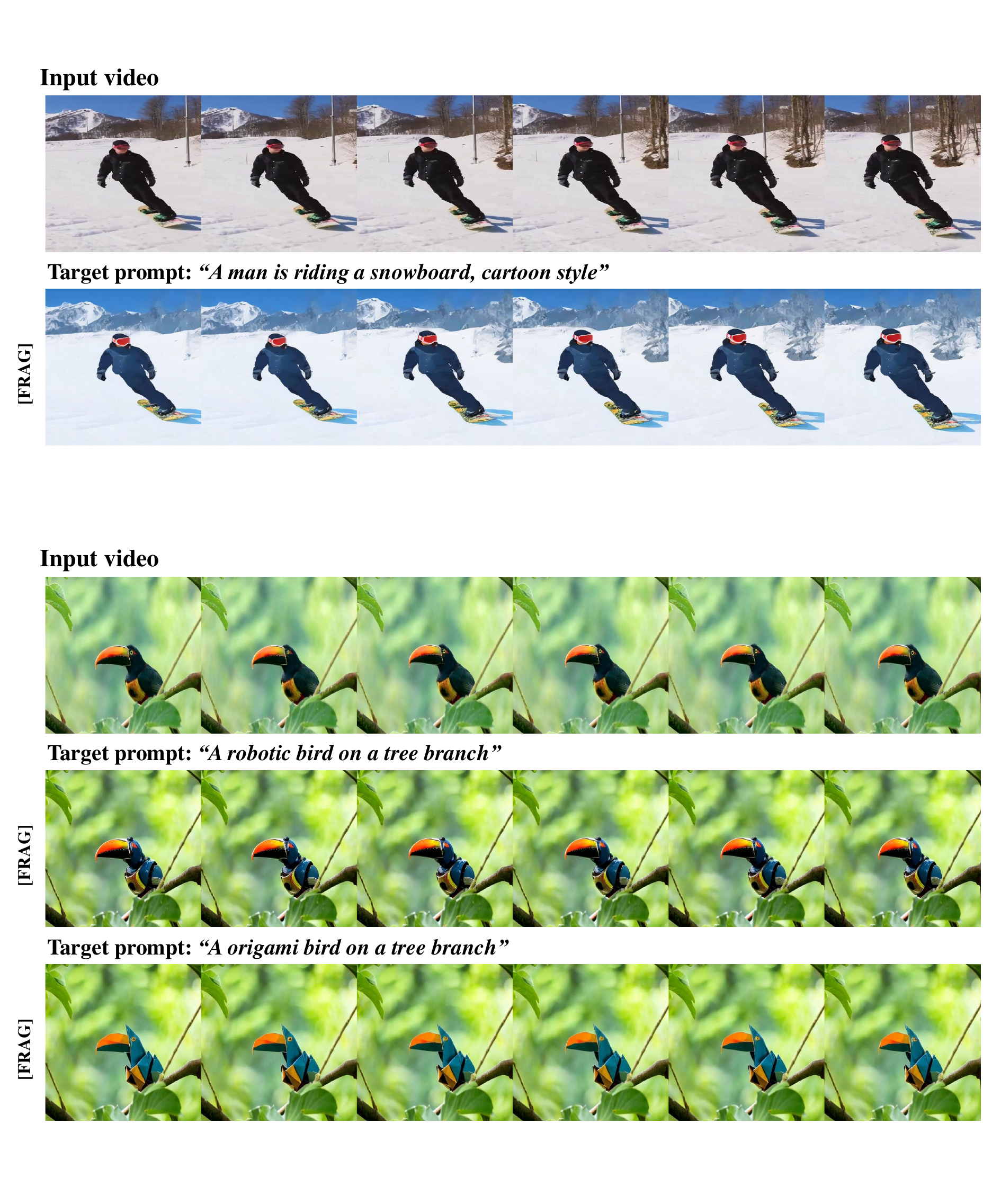}
\label{fig:page3}
\end{figure*}


\end{document}